\documentclass{article}

\usepackage{PRIMEarxiv}

\usepackage[utf8]{inputenc}     % allow utf-8 input
\usepackage[T1]{fontenc}        % use 8-bit T1 fonts
\usepackage{hyperref}           % hyperlinks
\usepackage{url}                % simple URL typesetting
\usepackage{booktabs}           % professional-quality tables
\usepackage{amsmath}            % math environments
\usepackage{amssymb}            % math symbols
\usepackage{amsfonts}           % blackboard math symbols
\usepackage{amsthm}             % theorem environments
\usepackage{nicefrac}           % compact symbols for 1/2, etc.
\usepackage{microtype}          % microtypography
\usepackage{graphicx}           % graphics
\usepackage[small]{caption}     % caption formatting
\usepackage{fancyhdr}           % header
\usepackage{soul}               % text highlighting
\usepackage{algorithm}          % algorithm environment
\usepackage{algorithmic}        % algorithmic typesetting
\usepackage[dvipsnames, svgnames, x11names]{xcolor}  % colors
\usepackage{colortbl}           % colored tables
\usepackage{multirow}           % multirow cells in tables
\usepackage{seqsplit}           % split long sequences
\usepackage[switch]{lineno}     % line numbers
\usepackage{lipsum}             % dummy text
\graphicspath{{media/}}     % organize your images and other figures under media/ folder

\title{DeepMoLM: Leveraging Visual and Geometric Structural Information for Molecule-Text Modeling}

\author{
  Jing Lan$^{1}$\thanks{Co-first author}, Hexiao Ding$^{1*}$, Hongzhao Chen$^{1*}$, Yufeng Jiang$^{1}$, Nga-Chun Ng$^{1,2}$, \\
  \textbf{Gwing Kei Yip}$^{1,3}$, \textbf{Gerald W.Y. Cheng}$^{1}$, \textbf{Yunlin Mao}$^{1}$, \textbf{Jing Cai}$^{1}$, \textbf{Liang-ting Lin}$^{1}$, \textbf{Jung Sun Yoo}$^{1}$\thanks{Corresponding author} \\
  $^{1}$Department of Health Technology and Informatics, The Hong Kong Polytechnic University \\
  $^{2}$Department of Nuclear Medicine and PET, Hong Kong Sanatorium and Hospital \\
  $^{3}$Department of Diagnostic and Interventional Radiology, Queen Elizabeth Hospital \\
  Hong Kong SAR, China\\
  \texttt{\{jing-hti.lan, hexiao.ding, hongzhao.chen, yufeng.jiang, yunlin.mao\}@connect.polyu.hk} \\
  \texttt{\{wai-yeung.cheng, jing.cai, ltlin, jungsun.yoo\}@polyu.edu.hk} \\
  \texttt{sam.nc.ng@hksh.com, gwinky.yip@ha.org.hk} \\
}

\begin{document}
\maketitle

\begin{abstract}

AI models for drug discovery and chemical literature mining must interpret molecular images and generate outputs consistent with 3D geometry and stereochemistry. Most molecular language models rely on strings or graphs, while vision-language models often miss stereochemical details and struggle to map continuous 3D structures into discrete tokens. We propose \textbf{DeepMoLM}: \underline{Deep Mo}lecular \underline{L}anguage \underline{M}odeling, a dual-view framework that grounds high-resolution molecular images in geometric invariants derived from molecular conformations. DeepMoLM preserves high-frequency evidence from $1024 \times 1024$ inputs, encodes conformer neighborhoods as discrete Extended 3-Dimensional Fingerprints, and fuses visual and geometric streams with cross-attention, enabling physically grounded generation without atom coordinates. DeepMoLM improves PubChem captioning with a $12.3\%$ relative METEOR gain over the strongest generalist baseline while staying competitive with specialist methods. It produces valid numeric outputs for all property queries and attains MAE $13.64$~g/mol on Molecular Weight and $37.89$ on Complexity in the specialist setting. On ChEBI-20 description generation from images, it exceeds generalist baselines and matches state-of-the-art vision-language models. Code is available at \url{https://github.com/1anj/DeepMoLM}.
% Code is available at \url{https://anonymous.4open.science/r/DeepMoLM}.

%TODO

\end{abstract}

% keywords can be removed

\section{Introduction}

\begin{figure}[t]
    \centering
    \includegraphics[width=1\linewidth]{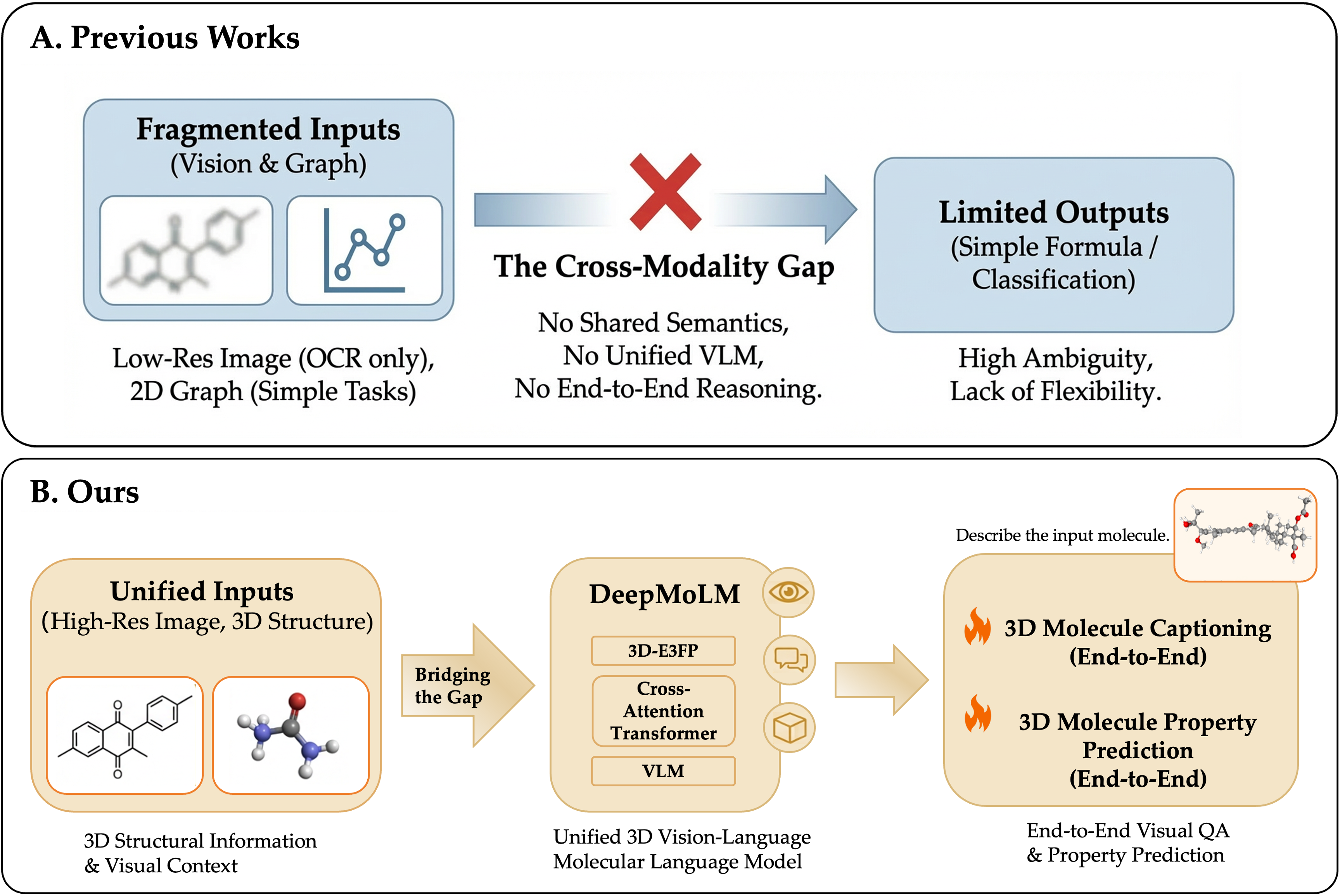}
    \caption{\textbf{Comparison of prior approaches with DeepMoLM.}
    \textbf{(A)} Previous studies represented molecules using low-resolution images processed through Optical Chemical Structure Recognition, or with 2D molecular graphs.
    \textbf{(B)} DeepMoLM integrates high-resolution molecular images with explicit 3D conformer-based structural descriptors. Through a cross-attention mechanism that links visual and geometric modalities within a Vision-Language Model, DeepMoLM supports end-to-end execution of complex tasks such as 3D molecule captioning and property prediction.}
    \label{fig:related_work}
\end{figure}

The physical nature of molecules follows three-dimensional (3D) geometric laws and quantum constraints~\cite{zhou2023unimol}, but much chemical knowledge is stored in scientific articles as molecular images, including 2D drawings and rendered structure figures~\cite{rajan2023decimer}. However, many AI models fail to align text with the underlying 3D geometry. As a result, these molecular images are often treated as useless illustrations rather than structured inputs. In this case, visual cues such as bond layout and stereochemical markers are not linked to a consistent physical conformation and the information in molecular images is not fully exploited.

% The physical nature of molecules is intrinsically governed by three-dimensional (3D) geometric laws and quantum constraints. However, due to the xxxx, most chemical knowledge is archived as two-dimensional visual depictions in scientific literature.
% From a data science perspective, this creates a fundamental feature misalignment: existing architectures decouple the modeling of spatial geometric priors (3D) from the processing of visual topological features (2D). While molecular images serve as a convenient, abundant low-dimensional proxy for structural data, they represent a lossy projection that explicitly encodes connectivity while implicitly discarding the conformational manifold.
% Conversely, while 3D conformers represent the ground truth that dictates function, they are computationally expensive to simulate and scarce in training corpora. Consequently, the critical research gap lies in the inability of current systems to invert this projection—that is, to infer latent 3D physical constraints from explicit 2D visual semantics. A robust molecular understanding system must therefore achieve dual-view consistency, reconciling the visual proxy with its underlying physical reality within a unified inference space.

Dual view consistency between molecular images and language hinges on three gaps. First, prevalent multimodal designs freeze unimodal encoders and rely on lightweight adapters for late fusion~\cite{liu2023molca}. Weak coupling encourages shortcut alignment, so the model tracks coarse semantics while neglecting geometric and stereochemical constraints. Second, projecting continuous 3D structure into a discrete language space can destroy geometric structure~\cite{li20243dmoleculetextinterpretationlanguage}. Physical invalidity occurs when rigid motion equivariance and stereochemical invariants are not preserved. Rotation or translation of the same molecule may alter the embedding. Enantiomers or closely related stereoisomers can then be indistinguishable.

Under conformational ambiguity, multiple realizations then collapse to similar representations, so fluent text is produced with incorrect chirality or spatial relations~\cite{fuchs2020se,alhamoud2024leveraging}. This is a geometric representation collapse rather than an energy-related issue. So, the third gap is clear. Molecular depictions encode semantics through localized high‑frequency cues such as stereobonds and ring closures. At standard resolutions, vision encoders often lose such details, and raising fidelity comes at a high cost since self‑attention grows quadratically~\cite{dosovitskiy2021imageworth16x16words}. DeepMoLM integrates image features with a 3D‑aware fingerprint branch through a fusion module, improving stereochemical stability under uncertain conformations and enabling cross‑modal reasoning without atom coordinates at inference.

We introduce DeepMoLM to unify molecular image understanding with 3D structure in an end-to-end vision-language framework. Within DeepMoLM, DeepEncoder is a dual-pathway module that preserves high-frequency cues in molecular images and enables efficient 1024×1024 processing through convolutional token compression. This architecture was originally developed for optical character recognition~\cite{wei2025deepseek}.
The local pathway captures fine stereochemical detail, and the global pathway provides structural coherence by aggregating long-range context. To bridge continuous 3D geometry with language models, we introduce explicit structural invariants using the Extended 3-Dimensional Fingerprint (E3FP)~\cite{axen2017simple}. Rather than concatenation, a cross-attention fusion projector makes visual tokens query geometric descriptors in order to ground image features before decoding.

The contributions are as follows. \textbf{(1)} We integrate a dual-pathway DeepEncoder that processes 1024$\times$1024 molecular images and preserves fine-grained stereochemical markers through convolutional token compression. \textbf{(2)} We introduce a cross-attention fusion projector that aligns visual representations with discrete 3D E3FP fingerprints, providing explicit geometric grounding within a unified multimodal embedding space. \textbf{(3)} Experimental results showed that the grounding in DeepMoLM improved performance on molecule captioning and property prediction, which are related to stereochemistry.

\section{Related Works}
\subsection{Molecular Representations} 

In drug discovery, molecular representation learning is often built on symbolic forms. A common approach represents molecules as linear strings, such as SMILES or SELFIES. 
Transformers are then pretrained on these strings to learn syntax, substructures, and generation patterns. MolT5 links molecule-to-text and text-to-molecule tasks, which supports transfer across multiple downstream settings~\cite{edwards2022translationmoleculesnaturallanguage}. Instruction tuning for chemistry and domain adaptation follow the same line, including ChemLLM and large scientific language models such as Galactica~\cite{pei20243dmolt5,zhang2024chemllm}. These methods still depend on 1D strings or 2D descriptors because tokenization and vocabularies in language models do not naturally encode full molecular structure. Graph neural networks address this challenge by modeling atoms and bonds as a 2D graph, where node states are iteratively updated through message passing~\cite{gilmer2017neural}. Self-supervised graph pretraining, such as GROVER, improves generalization in low-label settings by learning chemical regularities from large collections of unlabeled graphs~\cite{rong2020self}. Multi‑view pretraining adds geometric information to representation learning. GraphMVP aligns 2D topology with 3D conformer supervision to improve transferability~\cite{liu2021pretraining}.

\subsection{Vision Language Models for Molecular Images} 

Image-based methods add visual information to strengthen graph representations~\cite{xiang2024image}, and self-supervised training on molecular images shows 2D renderings can carry rich chemical contents~\cite{zeng2022accurate}. Optical Chemical Structure Recognition (OCSR) is designed to translate chemical figures from documents into graph representations interpretable by machines, including those in PDF and scanned formats~\cite{krasnov2024comparing}. MolScribe performs recognition through concurrent symbol parsing and molecular graph construction, ensuring connectivity is explicitly modeled in the decoding process~\cite{qian2023molscribe}. SwinOCSR and ChemVLM strengthen the visual backbone to process more complex inputs~\cite{xu2022swinocsr,li2025chemvlm}. Most OCSR methods still produce 2D graphs or strings without applying 3D geometric constraints. Moreover, stereochemistry remains uncertain when figures are noisy, of poor quality, or drawn in uncommon styles~\cite{krasnov2024comparing,qian2023molscribe}. OCSR digitizes explicit molecular structures, whereas vision-language models capture implicit semantics for higher-level reasoning. Molecular images encode information through atom labels, bond strokes, and stereochemical markers. These signals are sparse and localized rather than distributed across texture. Preserving them requires high resolution, but vision transformers are costly because self-attention scales quadratically with token count~\cite{dosovitskiy2021imageworth16x16words}. DeepEncoder employs a dual-pathway encoder with convolutional token compression and saves the token budget even while processing $1024 \times 1024$ molecular images~\cite{wei2025deepseek}.

\subsection{Geometric Learning and Multimodal Fusion} 

Methods such as Uni‑Mol and 3D models including SchNet, DimeNet, and GemNet embed inductive biases from physical geometry. They show that 3D invariants are important for quantum‑level property prediction~\cite{zhou2023unimol,schutt2017schnet,gasteiger2020directional,gasteiger2021gemnet}. SE(3) Transformer and EGNN enforce symmetry, allowing models to learn from conformers and molecular point clouds while maintaining invariance to spatial transformations~\cite{satorras2021en}. Recent work also links 3D encoders with language models for molecule–text interaction. 3D-MoLM aligns a 3D molecular encoder with a language model through a learned projector, which improves instruction following on 3D molecule–text data~\cite{li20243dmoleculetextinterpretationlanguage}. The method assumes access to structured 3D inputs and it does not target strong visual perception from molecular images. 3D‑MolT5 tokenizes fine‑grained 3D substructures and aligns them with sequence tokens in pretraining. This design strengthens the coupling between geometry and language~\cite{pei20243dmolt5}.
\section{Methodology}

\begin{figure*}[t]
    \centering
    \includegraphics[width=.85\linewidth]{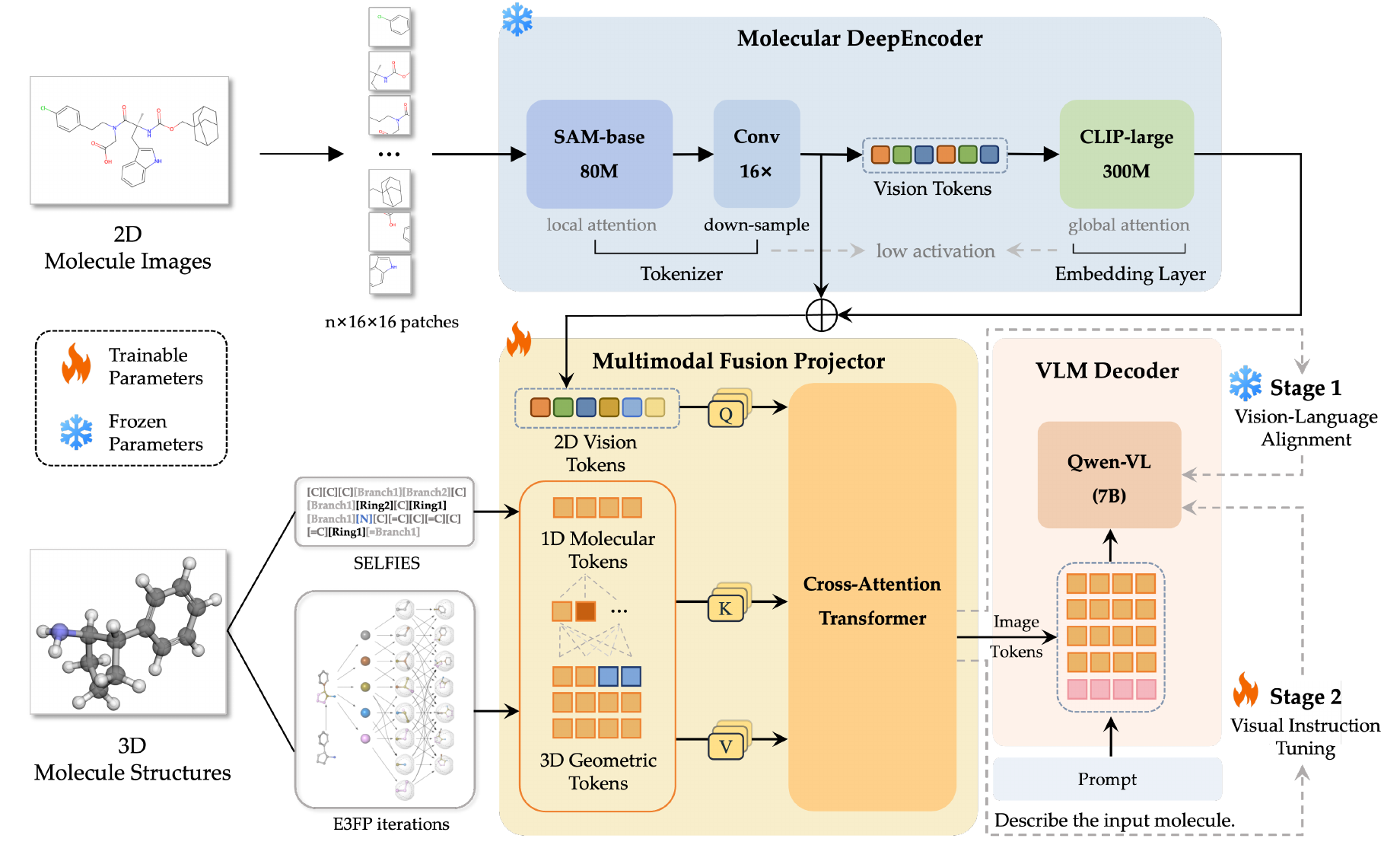}
    \caption{\textbf{Architecture of DeepMoLM.} The framework performs dual-view molecular understanding in three stages. (1) \textbf{DeepEncoder} extracts multi-scale features from a high-resolution molecular image using a SAM-Base local encoder, convolutional token compression, and a CLIP-Large global encoder. (2) \textbf{Fusion Projector} uses cross-attention to fuse 2D visual tokens with discrete 3D geometric fingerprints. (3) \textbf{VLM Decoder} generates captions and property text from the fused representation.}
    \label{fig:architecture}
\end{figure*}

The architecture of DeepMoLM is shown in Figure~\ref{fig:architecture}. It includes the \textbf{Molecular DeepEncoder} for high-resolution visual feature extraction, the \textbf{Multimodal Fusion Projector} for aligning visual and geometric modalities, and the \textbf{Vision-Language Decoder} built on Qwen2-VL~\cite{wang2024qwen2vlenhancingvisionlanguagemodels}.

Text generation is formulated as a multimodal autoregressive task. Each molecule $m$ is represented by a high-resolution image $\mathbf{I}_m \in \mathbb{R}^{H \times W \times C}$ with $H=W=1024$ and $C=3$, and by a structural token sequence $\mathbf{S}_m$ constructed from canonical SELFIES tokens and E3FP identifiers. Given a prompt $\mathbf{X}_{\mathrm{prompt}}$, the model generates the target sequence $\mathbf{y} = [y_1, \dots, y_T]$ by maximizing
\begin{equation}
    p(\mathbf{y} \mid \mathbf{X}_{\mathrm{prompt}}, \mathbf{I}_m, \mathbf{S}_m)
    = \prod_{t=1}^{T} p\!\left(y_t \mid \mathbf{y}_{<t}, \mathbf{X}_{\mathrm{prompt}}, \mathbf{I}_m, \mathbf{S}_m\right).
    \label{eq:prob_formulation}
\end{equation}

\subsection{Molecular Structure Representation}
\label{sec:molecule_fp}

\subsubsection{Structure-aware 1D Molecular Tokens}
We represent molecular topology with Self-referencing Embedded Strings (SELFIES)~\cite{Krenn_2020}. For molecule $m$, canonical SELFIES yields a token sequence
\begin{equation}
\mathcal{T}^{(1\mathrm{D})}_m = \{s_1, s_2, \dots, s_L\}.
\end{equation}
The length $L$ can exceed the number of heavy atoms due to non-atomic control tokens. We define an atom-position index set $\mathcal{A}\subseteq\{1,\dots,L\}$ and a bijection
\begin{equation}
\phi:\{1,\dots,N_{\mathrm{ha}}\}\rightarrow \mathcal{A},
\end{equation}
where $N_{\mathrm{ha}}$ denotes the number of heavy atoms. The mapping $\phi$ is obtained by parsing the canonical SELFIES into a molecular graph and recording the SELFIES positions that introduce heavy atoms.

\subsubsection{3D Geometric Tokens}

We represent conformer geometry using Extended 3-Dimensional Fingerprints (E3FP)~\cite{axen2017simple}. The conformer coordinates are denoted as $\mathbf{X}_m \in \mathbb{R}^{N_{\mathrm{ha}} \times 3}$. For each heavy atom $a_i$, E3FP generates hashed identifiers $\hat{d}_{i,j}$ across $K+1$ radii. The initial identifier $\hat{d}_{i,0}$ is derived from atom-level invariants. At iteration $j \in \{1,\dots,K\}$ with radius $R_j = r \cdot j$, E3FP collects neighbors within $R_j$ based on connectivity and stereochemical configuration from $\mathbf{X}_m$, then hashes the aggregate to obtain $\hat{d}_{i,j}$. We map each identifier into a discrete vocabulary of size $|F|$ by

\begin{equation}
d_{i,j} \triangleq \hat{d}_{i,j} \bmod |F|,
\qquad j\in\{0,\dots,K\}.
\end{equation}
This yields an atom-level 3D token tuple
\begin{equation}
\mathbf{d}_i = [d_{i,0}, d_{i,1}, \dots, d_{i,K}],
\qquad i\in\{1,\dots,N_{\mathrm{ha}}\},
\end{equation}
and an indexed 3D token table
\begin{equation}
\mathbf{D}_m \in \{0,\dots,|F|-1\}^{N_{\mathrm{ha}}\times (K+1)}, \quad \mathbf{D}_m[i,j] = d_{i,j}.
\end{equation}

SELFIES provides the discrete topological backbone $\mathcal{T}^{(1\mathrm{D})}_m$, and E3FP provides 3D descriptors indexed by heavy atoms that reflect molecular conformation. We align them by assigning $\mathbf{d}_i$ to the SELFIES position $t=\phi(i)$. Positions $t\notin\mathcal{A}$ have no aligned heavy atom and no associated 3D tuple.

\subsubsection{Token Embedding Fusion}
We use trainable embeddings
$\mathbf{E}_{\mathrm{1D}}:\mathcal{V}_{\mathrm{1D}}\rightarrow\mathbb{R}^{d_s}$
and
$\mathbf{E}_{\mathrm{3D}}:\{0,\dots,|F|-1\}\rightarrow\mathbb{R}^{d_s}$.
For each SELFIES position $t$, we compute
\begin{equation}
\mathbf{e}^{(1\mathrm{D})}_t=\mathbf{E}_{\mathrm{1D}}(s_t)\in\mathbb{R}^{d_s}.
\end{equation}
We define $m_t \triangleq \mathbb{I}[t\in\mathcal{A}]$. If $m_t=1$, let $i=\phi^{-1}(t)$ and compute
\begin{equation}
\mathbf{e}^{(3\mathrm{D})}_t
=\frac{1}{K+1}\sum_{j=0}^{K}\mathbf{E}_{\mathrm{3D}}(d_{i,j})
\in\mathbb{R}^{d_s}.
\end{equation}
If $m_t=0$, we set $\mathbf{e}^{(3\mathrm{D})}_t=\mathbf{0}\in\mathbb{R}^{d_s}$. We fuse 1D and 3D embeddings by
\begin{equation}
\label{eq:e3fp}
\mathbf{s}_t
=\frac{\mathbf{e}^{(1\mathrm{D})}_t+m_t\,\mathbf{e}^{(3\mathrm{D})}_t}{1+m_t}
\in\mathbb{R}^{d_s}.
\end{equation}
The structural sequence is $\mathbf{S}_m=[\mathbf{s}_1,\dots,\mathbf{s}_L]$, which serves as input to the multimodal fusion projector.

\subsection{Molecular DeepEncoder}
Molecular images contain high-frequency cues that determine chemical meaning, including atom glyphs, bond orders, ring closures, and stereochemical wedges. We encode $\mathbf{I}_m \in \mathbb{R}^{1024 \times 1024 \times 3}$ with a hierarchical encoder composed of a SAM-Base local vision transformer~\cite{kirillov2023segment}, a convolutional token compressor, and a CLIP-Large global vision transformer~\cite{radford2021learningtransferablevisualmodels}.

\subsubsection{Local Window Vision Transformer}
We divide $\mathbf{I}_m$ into non-overlapping patches of size $p=16$, giving a token sequence length $N=(1024/p)^2=4096$. 
A 12-layer SAM-Base vision transformer encodes these tokens with window attention of size $w=14$. 
The output is
\begin{equation}
\mathbf{H}_{\mathrm{local}} \in \mathbb{R}^{N \times d}, 
\quad N=4096,\quad d=1024.
\end{equation}

\subsubsection{Convolutional Token Compressor}
We reshape $\mathbf{H}_{\mathrm{local}}$ into a grid $\tilde{\mathbf{H}}_{\mathrm{local}} \in \mathbb{R}^{64 \times 64 \times d}$. We apply two $3 \times 3$ convolutions with stride $2$ and padding $1$, which downsample the grid to $16 \times 16$ while keeping the channel dimension $d$. The compressed sequence is
\begin{equation}
\mathbf{H}_{\mathrm{cmp}} \triangleq \operatorname{Compress}(\mathbf{H}_{\mathrm{local}}) \in \mathbb{R}^{M \times d},~ M=256.
\end{equation}

\subsubsection{Global Attention Transformer}
A 24-layer CLIP-Large vision transformer applies dense global self-attention to $\mathbf{H}_{\mathrm{cmp}}$ and outputs
\begin{equation}
\mathbf{H}_{\mathrm{global}} \in \mathbb{R}^{M \times d},
\quad M=256,\quad d=1024.
\end{equation}
At this stage, long‑range dependencies and structural motifs such as rings and functional groups are identified.

\subsubsection{Local and Global Feature Fusion}
We concatenate the global tokens with the local tokens along the channel dimension
\begin{equation}
\mathbf{H}_{\mathrm{vis}}
= \operatorname{Concat}\!\left(\mathbf{H}_{\mathrm{global}}, \mathbf{H}_{\mathrm{local}}\right)
\in \mathbb{R}^{M \times 2d}
= \mathbb{R}^{256 \times 2048}.
\end{equation}
The fused representation $\mathbf{H}_{\mathrm{vis}}$ is consumed by the multimodal fusion projector.

\subsection{Multimodal Fusion Projector}
\label{sec:fusion_projector}

The Multimodal Fusion Projector adds 3D structural information to the 2D visual token stream before the vision‑language decoder. It is a post‑normalization Transformer block with cross‑attention, using visual tokens as queries and structural tokens as keys and values. The DeepEncoder outputs $\mathbf{H}_{\mathrm{vis}}\in\mathbb{R}^{N_v\times d_v}$ with $N_v=256$ and $d_v=2048$. The molecular structural sequence from Section~\ref{sec:molecule_fp} is denoted by $\mathbf{S}_m\in\mathbb{R}^{L\times d_s}$, where $L$ is the SELFIES length and $d_s$ is the structural embedding dimension. To support batching, we pad the structural sequence to length $L_{\max}$ and write $\mathbf{S}_m^{\mathrm{pad}}\in\mathbb{R}^{L_{\max}\times d_s}$.

The fusion process is not autoregressive, so no causal mask is applied. We project both modalities into a shared hidden dimension $d_h$ that matches the decoder hidden size. The projected features are
\begin{equation}
\mathbf{H}_V\!=\!\mathbf{H}_{\mathrm{vis}}\mathbf{W}_V \!\in\! \mathbb{R}^{N_v\times d_h},
\quad
\mathbf{H}_S\!=\!\mathbf{S}_m^{\mathrm{pad}}\mathbf{W}_S \!\in\! \mathbb{R}^{L_{\max}\times d_h},
\end{equation}
with learnable $\mathbf{W}_V\in\mathbb{R}^{d_v\times d_h}$ and $\mathbf{W}_S\in\mathbb{R}^{d_s\times d_h}$.

Following the procedure in Algorithm \ref{alg:deepencoder_e3fp}, an additive mask $\mathbf{M}$ is constructed to differentiate between valid data and padding. Specifically, we set $\mathbf{M}[:,t] = 0$ for actual structural positions and $\mathbf{M}[:,t] = -\infty$ for padding ($t > L$), ensuring the softmax operation ignores padded elements.

Let $N_h$ denote the number of heads and $d_k=d_h/N_h$ the per-head dimension. For each head $h\in\{1,\dots,N_h\}$, we compute
\begin{align}
\mathbf{Q}^{(h)} &= \mathbf{H}_V\mathbf{W}_{Q}^{(h)} \in \mathbb{R}^{N_v \times d_k}, \\
\mathbf{K}^{(h)} &= \mathbf{H}_S\mathbf{W}_{K}^{(h)} \in \mathbb{R}^{L_{\max} \times d_k}, \\
\mathbf{V}^{(h)} &= \mathbf{H}_S\mathbf{W}_{V}^{(h)} \in \mathbb{R}^{L_{\max} \times d_k}.
\end{align}
where $\mathbf{W}_{Q}^{(h)},\mathbf{W}_{K}^{(h)},\mathbf{W}_{V}^{(h)}\in\mathbb{R}^{d_h\times d_k}$ are learnable parameters. The attention weights and head outputs are
\begin{align}
\mathbf{A}^{(h)} &=
\operatorname{Softmax}_{\mathrm{last}}\!\left(\frac{\mathbf{Q}^{(h)}\mathbf{K}^{(h)\top}}{\sqrt{d_k}}+\mathbf{M}\right)
\in\mathbb{R}^{N_v\times L_{\max}},\\
\mathbf{O}^{(h)} &= \mathbf{A}^{(h)}\mathbf{V}^{(h)} \in\mathbb{R}^{N_v\times d_k}.
\end{align}
The multi-head cross-attention output is
\begin{equation}
\operatorname{MHA}_{\mathrm{cross}}(\mathbf{H}_V,\mathbf{H}_S)
=
\operatorname{Concat}\!\left(\mathbf{O}^{(1)},\dots,\mathbf{O}^{(N_h)}\right)\mathbf{W}_O,
\end{equation}
with $\mathbf{W}_O\in\mathbb{R}^{d_h\times d_h}$.

We use residual connections with post-normalization. The cross-attention update is
\begin{equation}
\mathbf{H}_{\mathrm{cross}}
=
\operatorname{LayerNorm}\!\left(\mathbf{H}_V+\operatorname{MHA}_{\mathrm{cross}}(\mathbf{H}_V,\mathbf{H}_S)\right).
\end{equation}
A position-wise feed-forward network then refines the representation
\begin{equation}
\operatorname{FFN}(\mathbf{H}_{\mathrm{cross}})
=
\operatorname{GELU}\!\left(\mathbf{H}_{\mathrm{cross}}\mathbf{W}_1+\mathbf{b}_1\right)\mathbf{W}_2+\mathbf{b}_2,
\end{equation}
where $\mathbf{W}_1\in\mathbb{R}^{d_h\times d_{\mathrm{ff}}}$ and $\mathbf{W}_2\in\mathbb{R}^{d_{\mathrm{ff}}\times d_h}$. The fused output is
\begin{equation}
\mathbf{H}_{\mathrm{fused}}
=
\operatorname{LayerNorm}\!\left(\mathbf{H}_{\mathrm{cross}}+\operatorname{FFN}(\mathbf{H}_{\mathrm{cross}})\right)
\in\mathbb{R}^{N_v\times d_h}.
\end{equation}
For the 7B decoder, $d_h=4096$ and $N_v=256$. The output preserves the visual token budget while conditioning each visual token on structural descriptors derived from molecular conformations through cross-attention.

\begin{algorithm}[tb]
\caption{3D Tokenization and Sequence Padding}
\label{alg:deepencoder_e3fp}
\textbf{Input:} Molecule $m$ (atoms $\{a_i\}_{i=1}^{N_{\mathrm{ha}}}$), SELFIES length $L$\\
\textbf{Hyperparams:} E3FP iterations $K$, radius $r$, hash space $|F|$, max len $L_{\max}$\\
\textbf{Output:} Structural tokens $\mathbf{S}_m$, padding mask $\mathbf{M}$\\
    \vspace{-1em}
    \begin{algorithmic}[1]
    \FOR{$i=1$ \TO $N_{\mathrm{ha}}$} 
        \STATE $A_i \gets \mathrm{AtomicInvariants}(a_i)$
        \STATE $\hat{d}_{i,0} \gets \mathrm{MurmurHash3}(A_i)$
    \ENDFOR

    \FOR{$j=1$ \TO $K$}
        \STATE $R_j \gets r \cdot j$
        \FOR{$i=1$ \TO $N_{\mathrm{ha}}$}
            \STATE $\mathcal{N}_{i,j} \gets \emptyset$
            \FOR{$k=1$ \TO $N_{\mathrm{ha}}$}
                \IF{$\lVert \mathbf{x}_k - \mathbf{x}_i \rVert_2 \le R_j$}
                    \STATE $c_{k}^{i} \gets \mathrm{Connectivity}(a_k,a_i)$
                    \STATE $\sigma_{k}^{i} \gets \mathrm{Stereochemistry}(a_k,a_i)$
                    \STATE $\mathcal{N}_{i,j} \gets \mathcal{N}_{i,j} \cup \{(c_{k}^{i},\hat{d}_{k,j-1},\sigma_{k}^{i})\}$
                \ENDIF
            \ENDFOR
            \STATE $L_{i,j} \gets [\,j,\hat{d}_{i,j-1}\,] \oplus \mathrm{Sort}(\mathcal{N}_{i,j})$
            \STATE $\hat{d}_{i,j} \gets \mathrm{MurmurHash3}(L_{i,j})$
        \ENDFOR
    \ENDFOR

    \FOR{$i=1$ \TO $N_{\mathrm{ha}}$}
        \STATE $\mathbf{d}_i \gets [\hat{d}_{i,0},\hat{d}_{i,1},\dots,\hat{d}_{i,K}] \bmod |F|$
    \ENDFOR

    \STATE $\mathbf{S}_m \gets \mathbf{0}\in\mathbb{R}^{L_{\max}\times d_s}$
    \FOR{$t=1$ \TO $L_{\max}$}
        \STATE \textit{Generate $\mathbf{s}_t$ by Eq. (\ref{eq:e3fp}) and set $\mathbf{S}_m[t] \gets \mathbf{s}_t$}
        \STATE $\mathbf{M}[:,t] \gets 0$ if $t \le L$ else $-\infty$
    \ENDFOR
\end{algorithmic}
\end{algorithm}

\subsection{Vision Language Model (VLM) Decoder}
We use Qwen2-VL as the vision-language decoder. The decoder consumes the fused visual tokens $\mathbf{H}_{\mathrm{fused}}\in\mathbb{R}^{N_v\times d_h}$ with $N_v=256$ and the prompt token sequence $\mathbf{X}_{\mathrm{prompt}}=\{x_1,\dots,x_P\}$. Let $\mathbf{E}_{\mathrm{txt}}(\cdot)$ denote the token embedding lookup of Qwen2-VL, producing embeddings in $\mathbb{R}^{d_h}$. The embedded prompt is $\mathbf{H}_{\mathrm{txt}}=[\mathbf{E}_{\mathrm{txt}}(x_1),\dots,\mathbf{E}_{\mathrm{txt}}(x_P)]\in\mathbb{R}^{P\times d_h}$. The concatenated context is
\begin{equation}
\mathbf{H}_{\mathrm{in}} = [\mathbf{H}_{\mathrm{fused}};\mathbf{H}_{\mathrm{txt}}]\in\mathbb{R}^{(N_v+P)\times d_h}.
\end{equation}
The decoder generates the output token sequence $\mathbf{y}=[y_1,\dots,y_T]$ autoregressively. For step $t$, let $\mathbf{y}_{<t}$ be the prefix. The model computes
\begin{gather}
\mathbf{h}_t = f_{\mathrm{vlm}}(\mathbf{H}_{\mathrm{in}}, \mathbf{y}_{<t}) \in \mathbb{R}^{d_h}, \label{eq:vlm_decoding_1} \\
\boldsymbol{\pi}_t = \operatorname{Softmax}\!\left(f_{\mathrm{vocab}}(\mathbf{h}_t)\right)\in\mathbb{R}^{|\mathcal{V}|}, \label{eq:vlm_decoding_2}
\end{gather}
where $f_{\mathrm{vocab}}:\mathbb{R}^{d_h}\rightarrow\mathbb{R}^{|\mathcal{V}|}$ is the vocabulary projection. The conditional probability is $p(y_t=w\mid \mathbf{H}_{\mathrm{in}},\mathbf{y}_{<t})=\boldsymbol{\pi}_t[w]$ for $w\in\mathcal{V}$. In evaluation, we decode by
\begin{equation}
\tilde{y}_t = \arg\max_{w\in\mathcal{V}} \ p(y_t=w\mid \mathbf{H}_{\mathrm{in}},\mathbf{y}_{<t}),
\end{equation}
or by sampling from $\boldsymbol{\pi}_t$.

\subsection{Training Strategy}
We use a two-stage training pipeline. Stage-1 aligns the Multimodal Fusion Projector with a frozen decoder, so that the fused representation $\mathbf{H}_{\mathrm{fused}}$ serves as a compatible prefix. Stage-2 applies multimodal instruction tuning to adapt the decoder to molecular prompts conditioned on $\mathbf{H}_{\mathrm{fused}}$. Both stages optimize the same autoregressive negative log-likelihood objective in Equation~\ref{eq:prob_formulation}.

\subsubsection{Stage-1 Pre-training}
\paragraph{Vision-Language Alignment} We freeze the DeepEncoder and all decoder parameters, and train only the Multimodal Fusion Projector on molecular image–text pairs.

\subsubsection{Stage-2 Fine-tuning}
\paragraph{Visual Instruction Tuning}
Starting from the Stage-1 checkpoint, we keep the DeepEncoder frozen and jointly optimize the Multimodal Fusion Projector and the decoder on instruction style data such as captioning and property prediction, enabling the model to generate task-specific textual outputs conditioned on $\mathbf{H}_{\mathrm{fused}}$.

\section{Experiments}

\subsection{Implementation Details}
We used the two-stage training pipeline described in the training strategy and reported implementation details and hyperparameters for reproducibility. All models were implemented in PyTorch and trained on a single node with 8 NVIDIA H800 GPUs (80GB PCIe). Unless noted otherwise, the decoder backbone was \textit{Qwen2-VL-7B-Instruct}, and the molecular image encoder was a pretrained \textit{SAM-CLIP} Molecular DeepEncoder.

\paragraph{Stage-1 Pre-training}
We trained for 5 epochs with global batch size 128 (16 per GPU), using AdamW~\cite{loshchilov2019decoupledweightdecayregularization} with learning rate $10^{-4}$, cosine decay, and warmup ratio 0.03. Training used BF16 and a maximum context length of 4096.

\paragraph{Stage-2 Fine-tuning}
Unless otherwise specified, Stage-2 fine-tuning initialized from the Stage-1 checkpoint and updated the decoder backbone and the multimodal fusion projector while keeping the molecular image encoder frozen. We used DeepSpeed ZeRO-3~\cite{rajbhandari2020zeromemoryoptimizationstraining} for full-parameter fine-tuning on eight H800 GPUs, and trained for 10 epochs with a maximum learning rate of $5\times 10^{-5}$ and a global batch size of 64. We reported two variants. The Generalist model was trained with a unified multi-task objective over a mixture of all downstream datasets, while the Specialist model was fine-tuned separately for each downstream task using task-specific supervision with a warmup ratio of 0.01.

% \begin{table}[t]
%     \centering
%     \caption{\textbf{Performance on Molecule Captioning Task} The best results are highlighted in \textbf{bold}.}
%     \label{tab:captioning}
%     \resizebox{\linewidth}{!}{
%     \begin{tabular}{lccccc}
%         \toprule
%         \textbf{Model} & \textbf{Modality} & \textbf{BLEU-2} & \textbf{BLEU-4} & \textbf{ROUGE-L} & \textbf{METEOR} \\
%         \midrule
%         MolT5-Large~\cite{Edwards2022-molt5} & 1D & 42.8 & 31.2 & 52.1 & 51.3 \\
%         BioT5~\cite{Pei2023-biot5} & 1D & 46.2 & 34.5 & 55.8 & 54.1 \\
%         \midrule
%         MoMu~\cite{Su2022-momu} & 2D Graph & 48.3 & 36.1 & 57.2 & 55.9 \\
%         GIT-Mol~\cite{Liu2023-gitmol} & 2D Image & 52.4 & 40.8 & 60.1 & 57.8 \\
%         \midrule
%         3D-MoLM~\cite{Li2023-3dmolm} & 3D & 55.1 & 44.2 & 62.9 & 59.3 \\
%         % \textbf{DeepMoLM (Ours)} & 2D+3D & \textbf{57.8} & \textbf{47.5} & \textbf{65.4} & \textbf{61.4} \\
%         \bottomrule
%     \end{tabular}
%     }
% \end{table}

% \subsection{Experimental Results}

\subsection{Molecule Captioning}
We evaluated DeepMoLM on molecule captioning to measure its understanding of three-dimensional molecular structure. We used the PubChem dataset~\cite{pubchem2021}, which contained about 15,000 pairs of 3D molecular structures and PubChem textual descriptions. Each caption included molecular names and 3D conformation-related statements. This setting jointly evaluated name prediction following Favre and Powell and description prediction following Edwards~\cite{edwards2022translationmoleculesnaturallanguage}, providing a stronger test than property description alone.

We compared against baselines from two groups. Specialist models were optimized for molecular tasks, including MolT5-Large~\cite{edwards2022translationmoleculesnaturallanguage}, MoMu-Large~\cite{su2022molecularmultimodalfoundationmodel}, UniMoT~\cite{guo2025unimotunifiedmoleculetextlanguage}, and 3D-MoLM~\cite{li20243dmoleculetextinterpretationlanguage}, where we also reported a variant pre-trained without GPT-3.5 enrichment. Generalist models targeted broader multimodal generation or relied on large language models, including 2D-MoLM~\cite{li20243dmoleculetextinterpretationlanguage}, Llama2-7B~\cite{touvron2023llama2openfoundation}, and Qwen2-VL-7B~\cite{wang2024qwen2vlenhancingvisionlanguagemodels}. We reported BLEU-2, BLEU-4, ROUGE-1, ROUGE-2, ROUGE-L, and METEOR.

Table~\ref{tab:molecule_captioning} showed that DeepMoLM was competitive with both specialist and generalist baselines. Among specialist models, it matched the strong performance of UniMoT, which relied on a pretrained molecular encoder. Although DeepMoLM used only molecular images, it captured fine-grained structural cues and achieved higher ROUGE-1 and METEOR scores than specialist baselines. Among generalist models, DeepMoLM clearly outperformed standard LLMs and vision-language models, yielding more than 10\% relative gain in METEOR over the second-best baseline. Overall, DeepMoLM aligned visual molecular representations with natural language and generated informative captions without explicit atom coordinates.

\begin{table}[h]
    \centering
    \caption{Peformance on Molecule Captioning. $\dagger$ refers to a variant of 3D-MoLM that is initially pre-trained on the original PubChem text without GPT-3.5 enrichment. Best results are \textbf{bolded}, second best are \underline{underlined}.}
    \label{tab:molecule_captioning}
    \resizebox{\columnwidth}{!}{%
    \begin{tabular}{llcccccc}
        \toprule
        \textsc{Type} & \textsc{Model} & \textsc{BLEU-2} & \textsc{BLEU-4} & \textsc{ROUGE-1} & \textsc{ROUGE-2} & \textsc{ROUGE-L} & \textsc{METEOR} \\
        \midrule
        \multirow{8}{*}{\textbf{\textit{Specialist}}} 
        & MolT5-Large & 25.87 & 17.28 & 34.07 & 16.42 & 23.41 & 28.04 \\
        & MoMu-Large & 26.34 & 18.01 & 34.75 & 16.86 & 24.76 & 28.73 \\
        & 3D-MoLM$\dagger$ & 29.82 & 22.39 & 37.23 & 22.49 & 31.07 & 32.69 \\
        & 3D-MoLM & 30.32 & 22.52 & 36.84 & 22.32 & \underline{31.23} & 33.06 \\
        & UniMoT & \underline{31.30} & \underline{23.80} & \underline{37.50} & \textbf{23.70} & \textbf{33.60} & \underline{34.80} \\
        & \cellcolor{gray!20}\textbf{DeepMoLM} & \cellcolor{gray!20}\textbf{33.78} & \cellcolor{gray!20}\textbf{25.53} & \cellcolor{gray!20}\textbf{44.49} & \cellcolor{gray!20}\underline{22.57} & \cellcolor{gray!20}29.04 & \cellcolor{gray!20}\textbf{35.87} \\
        \midrule
        \multirow{6}{*}{\textbf{\textit{Generalist}}} 
        & 2D-MoLM & 27.15 & 21.19 & 36.02 & 20.76 & 29.12 & 32.28 \\
        & 3D-MoLM$\dagger$ & \underline{29.25} & \underline{22.07} & 36.48 & \underline{21.80} & \textbf{30.95} & 33.12 \\
        & 3D-MoLM & 28.95 & 21.63 & \underline{36.51} & 21.26 & 30.02 & \underline{33.55} \\
        & Llama2-7B & 27.01 & 20.94 & 35.76 & 20.68 & 28.88 & 32.11 \\
        & Qwen2-VL-7B & 18.91 & 8.33 & 36.38 & 10.90 & 19.55 & 23.44 \\
        & \cellcolor{gray!20}\textbf{DeepMoLM} & \cellcolor{gray!20}\textbf{35.87} & \cellcolor{gray!20}\textbf{25.53} & \cellcolor{gray!20}\textbf{47.07} & \cellcolor{gray!20}\textbf{23.07} & \cellcolor{gray!20}\underline{30.76} & \cellcolor{gray!20}\textbf{37.66} \\
        \bottomrule
    \end{tabular}
    }
\end{table}

\begin{table}[ht]
\centering
\caption{Performance on Computed Property Prediction. Mean Absolute Error (MAE) alongside the valid answer rate (in parentheses) are reported. $\dagger$ denotes a variant of 3D-MoLM pre-trained on original PubChem text without GPT-3.5 enrichment. * indicates the model was evaluated without fine-tuning. The best results are \textbf{bolded}, and the second best are \underline{underlined}.}
\label{tab:mae_results}
\resizebox{\columnwidth}{!}{% Resize table to fit text width if necessary
\begin{tabular}{llcccc}
\toprule
\textsc{Type} & \textsc{Model} & \textsc{Weight (g/mol)} $\downarrow$ & \textsc{LogP} $\downarrow$ & \textsc{TPSA (\AA$^2$)} $\downarrow$ & \textsc{Complexity} $\downarrow$ \\
\midrule
\textbf{\textit{Non-LM}} & Uni-Mol & 20.35 & \textbf{0.59} & 13.48 & 57.24 \\
\midrule
\multirow{6}{*}{\textbf{\textit{Specialist}}} 
& 2D-MoLM & 21.48 (94\%) & 0.88 (96\%) & 13.52 (92\%) & 55.74 (94\%) \\
& 3D-MoLM$\dagger$ & 16.18 (96\%) & 0.95 (96\%) & 10.26 (94\%) & 49.15 (95\%) \\
& 3D-MoLM & \underline{14.79} (95\%) & \textbf{0.66} (97\%) & \textbf{9.71} (93\%) & \underline{44.85} (94\%) \\
& Llama2-7B & 22.10 (96\%) & 1.45 (95\%) & 15.87 (92\%) & 69.74 (93\%) \\
& \cellcolor{gray!20}\textbf{DeepMoLM} & \cellcolor{gray!20}\textbf{13.64} (100\%) & \cellcolor{gray!20}\underline{0.73} (100\%) & \cellcolor{gray!20}\underline{10.04} (100\%) & \cellcolor{gray!20}\textbf{37.89} (100\%) \\
\midrule
\multirow{7}{*}{\textbf{\textit{Generalist}}} 
& 2D-MoLM & 20.80 (92\%) & \underline{1.36} (94\%) & 12.47 (89\%) & 52.70 (91\%) \\
& 3D-MoLM$\dagger$ & 19.54 (93\%) & 0.92 (92\%) & 11.14 (92\%) & 54.68 (90\%) \\
& 3D-MoLM & \underline{16.58} (92\%) & \textbf{0.78} (95\%) & \textbf{10.90} (90\%) & \underline{45.49} (89\%) \\
& Llama2-7B* & 42.18 (82\%) & 2.10 (85\%) & 27.11 (84\%) & 121.87 (76\%) \\
& Llama2-7B & 27.42 (92\%) & 1.78 (93\%) & 17.07 (90\%) & 78.16 (92\%) \\
& Qwen2-VL-7B & 42.13 (86\%) & 3.75 (99\%) & 50.89(32\%) & {103.07} (100\%) \\
& \cellcolor{gray!20}\textbf{DeepMoLM} & \cellcolor{gray!20}\textbf{14.63} (100\%) & \cellcolor{gray!20}\underline{1.36} (100\%) & \cellcolor{gray!20}\underline{11.14} (100\%) & \cellcolor{gray!20}\textbf{41.73} (100\%) \\
\bottomrule
\end{tabular}
}
\end{table}

\begin{table}[ht]
\centering
\caption{Performance on Molecule Description. Best results are \textbf{bolded}, second best are \underline{underlined}.}
\label{tab:molecule_description}
\resizebox{\columnwidth}{!}{% Resize table to fit text width if necessary
\begin{tabular}{llccccc}
\toprule
\textsc{METHOD} & \textsc{INPUT} & \textsc{BLEU-2} & \textsc{BLEU-4} & \textsc{ROUGE-2} & \textsc{ROUGE-L} & \textsc{METEOR}  \\
\midrule
\multicolumn{7}{l}{\textit{\textbf{Molecule Input (Specialist)}}} \\
\midrule
MoMu & SMILES & 54.9 & 46.2 & 47.9 & 57.5 & 57.6 \\
MolFM & SMILES & 58.5 & 49.8 & 50.8 & 59.4 & 60.7 \\
BioT5 & SMILES & \underline{63.5} & \underline{55.6} & \underline{55.9} & \underline{63.3} & \underline{65.6} \\
MolCA & SMILES & 62.0 & 53.1 & 53.7 & 61.8 & 65.1 \\
BioT5+  & SMILES & \textbf{66.6} & \textbf{59.1} & \textbf{58.4} & \textbf{65.0} & \textbf{68.1} \\
\midrule
\multicolumn{7}{l}{\textit{\textbf{Molecule Input (Generalist)}}} \\
\midrule
3D-MoLM & SMILES & 6.7 & 3.0 & 4.2 & 8.6 & 18.3 \\
Mol-Instructions & SMILES & 24.9 & 17.1 & 20.3 & 28.9 & 27.1  \\
BioMedGPT & SMILES & \underline{30.6} & \underline{19.8} & \underline{25.7} & \underline{38.3} & \underline{35.0} \\
GIT-Mol & SMILES & \textbf{35.2} & \textbf{26.3} & \textbf{48.5} & \textbf{56.0} & \textbf{43.0} \\
\midrule
\multicolumn{7}{l}{\textit{\textbf{Vision Input}}} \\
\midrule
MolScribe \& BioT5+ & Image & 53.43 & 44.64 & 46.48 & 55.30 & 57.09  \\
DoubleCheck \& BioT5+ & Image & 54.40 & 45.56 & 47.21 & 55.94 & 58.07 \\
Mol-VL-2B & Image & 50.57 & 40.26 & 41.99 & 52.25 & 53.34 \\
Mol-VL-7B & Image & \underline{55.73} & \underline{46.14} & \underline{47.26} & \textbf{56.61} & \underline{58.14}  \\
Qwen2-VL-7B & Image & 6.34 & 1.13 & 4.53 & 14.15 & 19.95  \\
\midrule
\rowcolor{gray!20} \textbf{DeepMoLM} & Image & \textbf{55.76} & \textbf{46.29} & \textbf{48.07} & \underline{55.96} & \textbf{58.23} \\
\bottomrule
\end{tabular}
}
\end{table}

\begin{table}[ht]
    \centering
    \caption{Case Studies for Molecule Captioning and Description.}
    \label{tab:case_study}
    \resizebox{\columnwidth}{!}{%
    \begin{tabular}{p{2cm} p{2cm} p{5.2cm} p{5.2cm}}
        \toprule
        \multicolumn{1}{c}{\textsc{Molecule}} & 
        \multicolumn{1}{c}{\textsc{Instruction}} & 
        \multicolumn{1}{c}{\textsc{DeepMoLM}} & 
        \multicolumn{1}{c}{\textsc{Ground Truth}} \\
        \midrule
        
        % Row 1 (保持不变)
        \raisebox{-4cm}{\includegraphics[width=2cm, height=2cm, keepaspectratio]{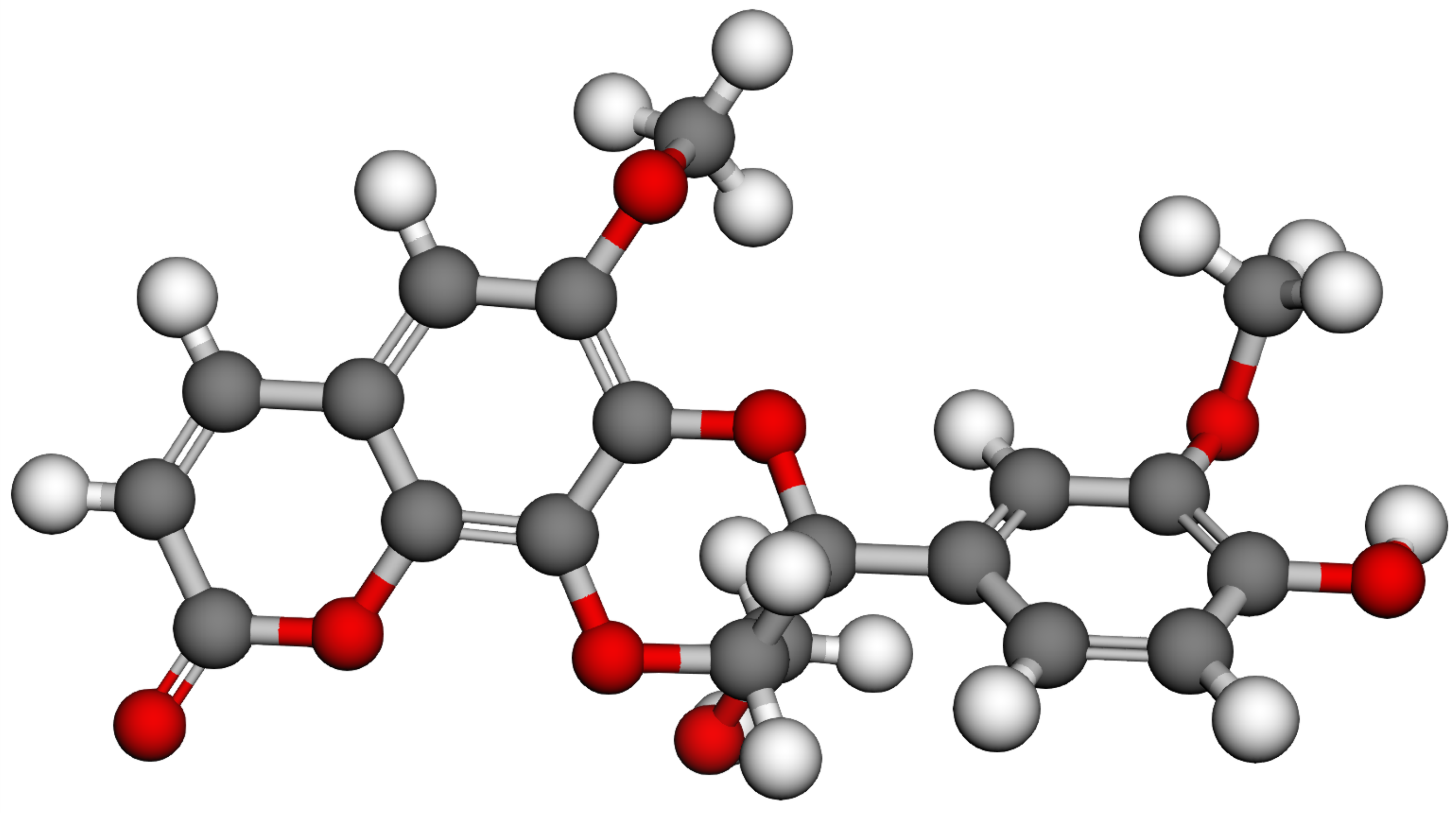}} 
        & 
        \multicolumn{1}{>{\raggedright\arraybackslash}p{2cm}}{Describe the input molecule.}
        & 
        \textcolor{Cerulean}{The molecule with the SMILES code \texttt{\seqsplit{COC1=C(C=CC(=C1)[C@@H]2C@@HCO)O}}} is a complex natural product known as \textcolor{Cerulean}{8'-epi-cleomiscosin A}, which is \textcolor{Cerulean}{isolated from the aerial parts of \textit{Rhododendron collettianum}}. It has a highly intricate structure consisting of multiple rings and functional groups. The molecule contains several hydroxyl and methoxy groups, which contribute to its relatively high polarity and ability to form hydrogen bonds. This property allows it to interact with other molecules through intermolecular forces, potentially influencing its solubility and reactivity. Additionally, the presence of oxygen-containing heterocycles and a lactone moiety suggests that this molecule may exhibit biological activity, as such structural features are common in bioactive natural products. The stereochemistry of the molecule is indicated by the presence of ``@'' symbols in the SMILES code, corresponding to a defined three-dimensional arrangement of atoms. This stereochemistry can affect its biological activity and interactions with enzymes or receptors. Notably, \textcolor{Cerulean}{8'-epi-cleomiscosin A} has been reported to show \textcolor{Cerulean}{inhibitory activity toward tyrosinase}, suggesting potential applications in \textcolor{Cerulean}{medicinal chemistry and cosmetic science}.
        & 
        \textcolor{Cerulean}{8'-epi-cleomiscosin A} is a complex organic molecule with \textcolor{Cerulean}{a SMILES code of \texttt{\seqsplit{COC1=C(C=CC(=C1)[C@@H]2C@@HCO)O}}}. It belongs to the phenols group and is categorized as an aromatic ether and an organic heterotricyclic compound. Structurally, \textcolor{Cerulean}{8'-epi-cleomiscosin A} is a substituted 2,3-dihydro-9H-[1,4]dioxino[2,3-h]chromen-9-one. It contains a 4-hydroxy-3-methoxy phenyl group at position 3, a hydroxymethyl group at position 2, and a methoxy group at position 5. Additionally, it is the 2S,3R stereoisomer. This molecule is \textcolor{Cerulean}{isolated from the aerial parts of \textit{Rhododendron collettianum}}. It has demonstrated \textcolor{Cerulean}{inhibitory activity towards} the enzyme \textcolor{Cerulean}{tyrosinase}, making it an EC 1.14.18.1 inhibitor. Tyrosinase is involved in melanin synthesis, so the inhibitory activity of \textcolor{Cerulean}{8'-epi-cleomiscosin A} suggests its potential in skin lightening or anti-pigmentation applications. In conclusion, \textcolor{Cerulean}{8'-epi-cleomiscosin A} is a unique organic compound with interesting chemical and physical properties. Its structure, as well as its inhibitory activity towards tyrosinase, make it a valuable molecule in the field of \textcolor{Cerulean}{medicinal chemistry and cosmetic science}. \\
        \midrule
        
        % Row 2 (保持不变)
        \raisebox{-2cm}{\includegraphics[width=2cm, height=2cm, keepaspectratio]{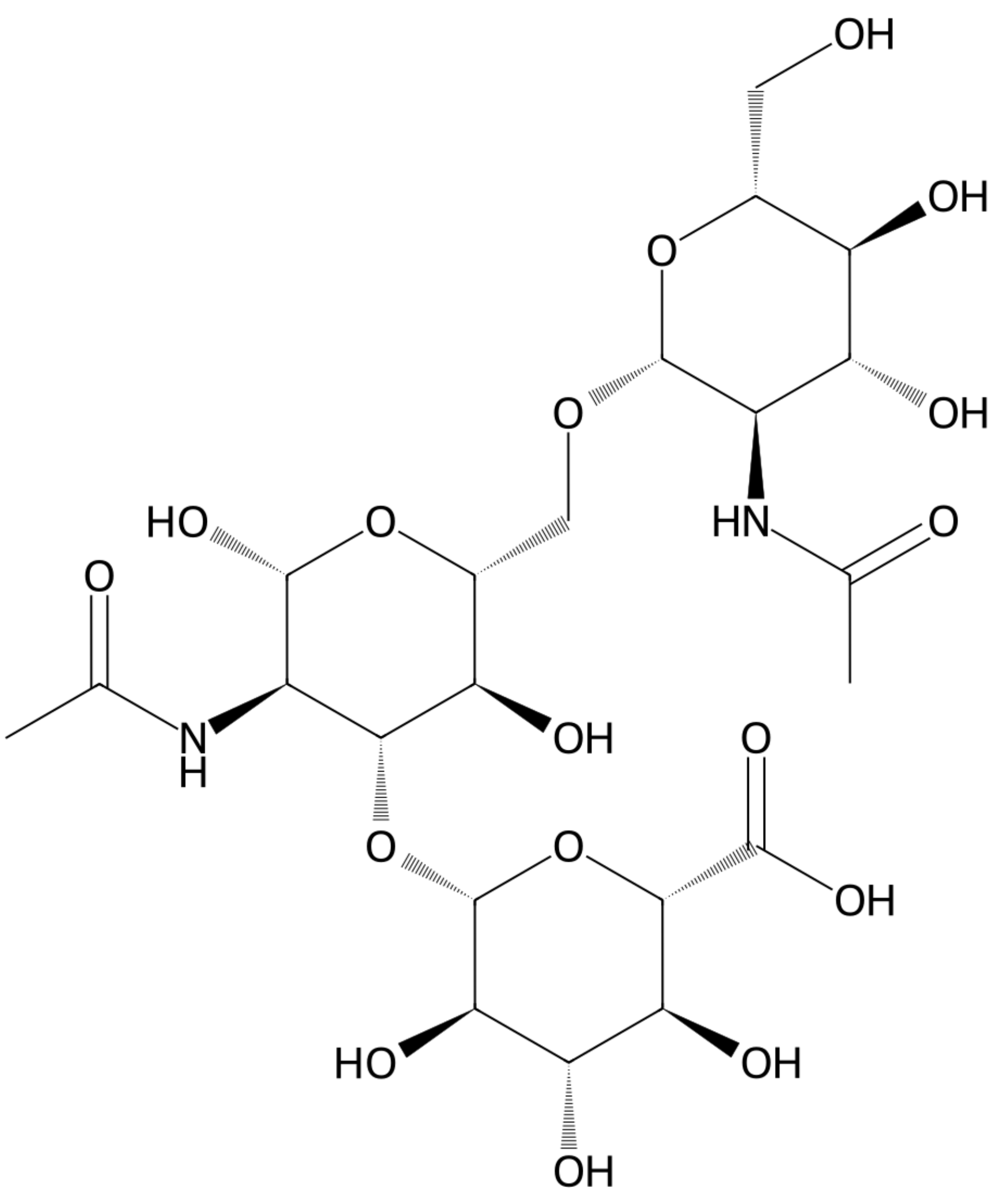}} 
        & 
        \multicolumn{1}{>{\raggedright\arraybackslash}p{2cm}}{Please describe this drug.}
        & 
        \textcolor{Cerulean}{The molecule is a branched amino} tetrasaccharide comprising \textcolor{Cerulean}{N-acetyl-beta-D-glucosamine} at the reducing end \textcolor{Cerulean}{having} a beta-D-galactosyl residue \textcolor{Cerulean}{attached at the} 6-position and a beta-D-galactosyl-(1$\to$3)-beta-D-galactosyl moiety attached at the 3-position.
        & 
        \textcolor{Cerulean}{The molecule is a branched amino} trisaccharide consisting of \textcolor{Cerulean}{N-acetyl-beta-D-glucosamine} \textcolor{Cerulean}{having} beta-D-glucuronosyl and N-acetyl-beta-D-glucosaminyl residues \textcolor{Cerulean}{attached at the} 3- and 6-positions respectively. It is an amino trisaccharide and a glucosamine oligosaccharide. \\

        % \midrule
        
        % % Row 3 (已修改：拆分为两行，使用 cmidrule 分割)
        % % 第一部分：图片 + 问题1 + 回答1
        % \multirow{2}{*}{\raisebox{-2cm}{\includegraphics[width=2cm, height=2cm, keepaspectratio]{section/img/000561_7_oxane.pdf}}}
        % & 
        % \multicolumn{1}{>{\raggedright\arraybackslash}p{2cm}}{What is the IUPAC of the molecule?}
        % & 
        % \multicolumn{2}{p{10.8cm}}{The IUPAC is (2S,3S,4S,5R,6R)-6-[(2R,3S,4R,5R,6R)-5-acetamido-2-[[(2R,3R,4R,5S,6R)-3-acetamido-4,5-dihydroxy-6-(hydroxymethyl)oxan-2-yl]oxymethyl]-3,6-dihydroxyoxan-4-yl]oxy-3,4,5-trihydroxyoxane-2-carboxylic acid.} \\
        
        % % 横线：贯穿第2列到第4列
        % \cmidrule{2-4}
        
        % % 第二部分：图片占位符(空) + 问题2 + 回答2
        %  & 
        % \multicolumn{1}{>{\raggedright\arraybackslash}p{2cm}}{Which functional group is highlighted?} 
        % & 
        % \multicolumn{2}{p{10.8cm}}{The functional group oxane is highlighted.} \\
        
        \bottomrule
    \end{tabular}%
    }
\end{table}

\begin{figure}[ht]
    \centering
    \includegraphics[width=.85\linewidth]{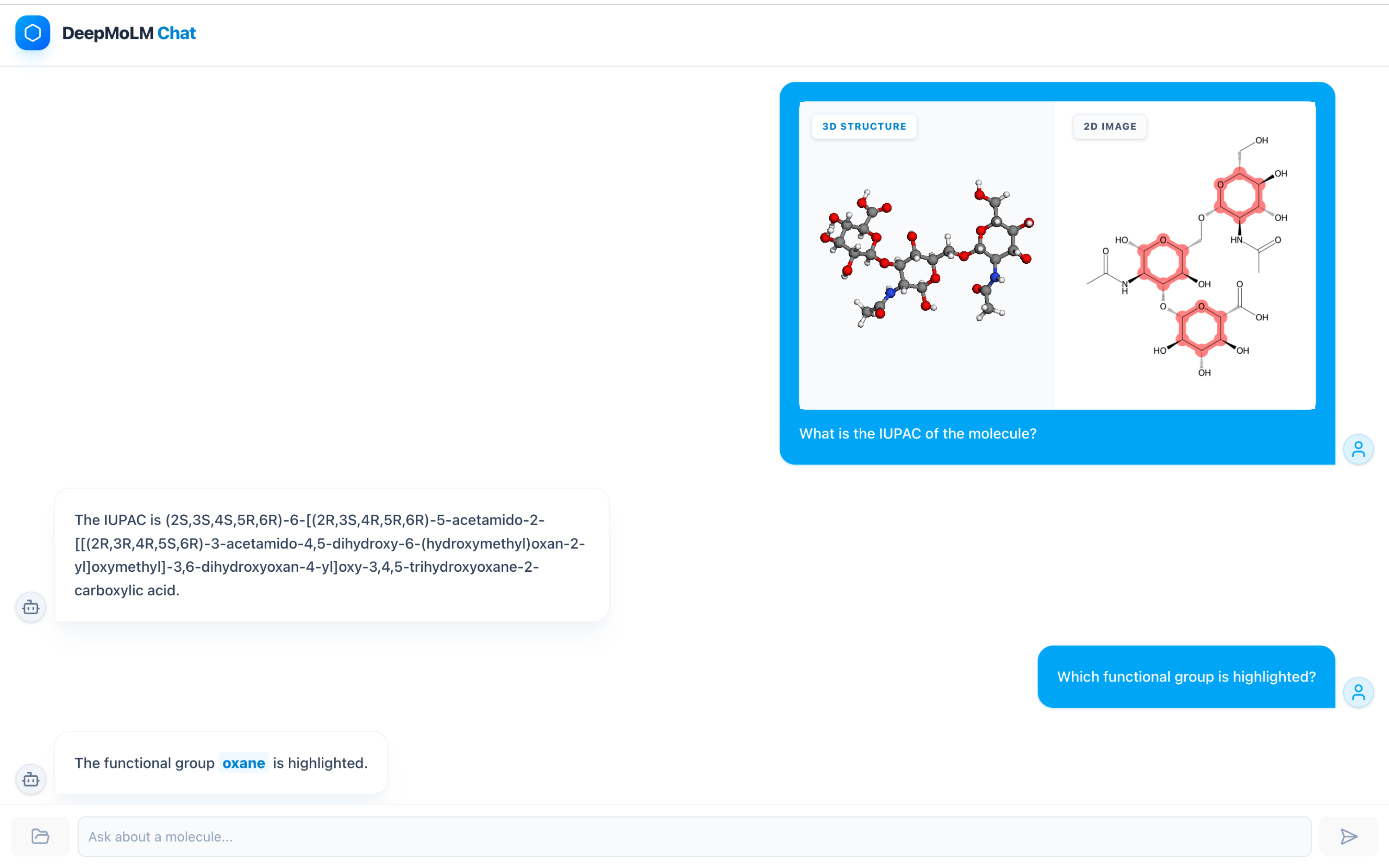}
    \caption{VLM Interface for Open-Text Molecular QA}
    \label{fig:gui}
\end{figure}

\begin{table}[htbp]
\centering
\textbf{}
\caption{Ablation Study. Results on Molecular Captioning (PubChem) and Molecular Description (ChEBI-20) for removing pre-training, the 3D-E3FP branch, or the fusion projector. \textbf{Bold} marks the best score. $\downarrow$ marks the largest drop in each metric.}
\label{tab:ablation_study}

\resizebox{\columnwidth}{!}{% Resize table to fit text width
\begin{tabular}{lcccccc}
\toprule
\textsc{Model} & \textsc{BLEU-2} & \textsc{BLEU-4} & \textsc{ROUGE-1} & \textsc{ROUGE-2} & \textsc{ROUGE-L} & \textsc{METEOR} \\
\midrule

% --- 修改部分 1: Molecular Captioning ---
\rowcolor{gray!20} 
\multicolumn{7}{l}{\textit{\textbf{Molecular Captioning}}} \\ 
\midrule
w/o Pre-training \& 3D-E3FP \& Fusion (linear) & 18.91 & 8.33 & 36.38~$\downarrow$ & 10.90~$\downarrow$ & 19.55~$\downarrow$ & 23.44 \\
DeepMoLM & \textbf{33.78} & \textbf{25.53} & \textbf{44.49} & \textbf{22.57} & \textbf{29.04} & \textbf{35.87} \\
\midrule

% --- 修改部分 2: Molecular Description ---
\rowcolor{gray!20}
\multicolumn{7}{l}{\textit{\textbf{Molecular Description}}} \\ 
\midrule
w/o Pre-training \& 3D-E3FP \& Fusion (linear) & 24.56 & 15.00 & 40.03~$\downarrow$ & 20.22~$\downarrow$ & 33.25~$\downarrow$ & 29.58 \\
w/o Pre-training \& Fusion (concat) & 19.94~$\downarrow$ & 12.96~$\downarrow$ & 42.33 & 22.18 & 36.67 & 27.29~$\downarrow$ \\
w/o Pre-training & 26.28 & 15.99 & 43.29 & 24.09 & 38.19 & 31.60 \\
DeepMoLM & \textbf{55.76} & \textbf{46.29} & \textbf{61.64} & \textbf{48.07} & \textbf{55.96} & \textbf{58.23} \\
\bottomrule
\end{tabular}
}
\end{table}

\subsection{Molecular Property Prediction}
We evaluated DeepMoLM on molecular property prediction using PubChem, covering Molecular Weight, LogP, TPSA, and Complexity. We compared with Uni-Mol, 2D-MoLM, 3D-MoLM, and generalist models including Llama2~\cite{touvron2023llama2openfoundation} and Qwen2-VL~\cite{wang2024qwen2vlenhancingvisionlanguagemodels}. DeepMoLM returned valid outputs for all queries, achieving a 100\% validity rate and avoiding the formatting failures frequently observed in generalist models. 

DeepMoLM matched or exceeded leading 3D approaches while using only 2D molecular images. Under the Specialist setting, it achieved the lowest MAE on Molecular Weight and Complexity, outperforming geometry-aware baselines including 3D-MoLM, and remained competitive on LogP and TPSA. This indicated that the visual encoder captured key structural cues without requiring explicit 3D coordinates. Under the Generalist setting, DeepMoLM substantially surpassed both vision-language and text-only models.

\subsection{Molecular Description}
Following OCSU~\cite{fan2025ocsu}, we converted CheBI-20~\cite{edwards2022translationmoleculesnaturallanguage} into an instruction-tuned format. To prevent leakage, we removed CheBI-20 test molecules that also appeared in the PubChem 3D molecule-text pairs used for pre-training. We further removed molecular names from the target text to avoid learning trivial name-to-sequence shortcuts. Baselines were taken from MolT5~\cite{edwards2022translationmoleculesnaturallanguage}, MolReGPT~\cite{Li_2024}, MolFM~\cite{luo2023molfmmultimodalmolecularfoundation}, GIT-Mol~\cite{Liu_2024}, MolXPT~\cite{liu2023molxptwrappingmoleculestext}, and BioT5~\cite{pei2024biot5enrichingcrossmodalintegration}.

Table~\ref{tab:molecule_description} reported results for molecule description generation. We compared DeepMoLM with specialist and generalist models that took SMILES as input, and with vision-language models that took images as input. DeepMoLM achieved the best results on most metrics and consistently surpassed all generalist baselines. Specialist models such as BioT5+ scored higher due to large-scale pre-training on 1D chemical sequences, yet DeepMoLM narrowed this gap while avoiding SMILES at inference. This suggested that DeepMoLM learned informative molecular representations directly from 2D images, which was useful when textual representations were missing or unreliable.

\subsection{Ablation Study}

To validate the effects of pre-training, the 3D molecular encoder, and multimodal fusion, we performed ablations on PubChem~\cite{pubchem2021} and ChEBI-20~\cite{edwards2022translationmoleculesnaturallanguage}. We compared DeepMoLM with three variants trained without pre-training. One was trained from scratch. Another replaced the fusion projector with feature concatenation. The third variant removed the 3D-E3FP branch and applied a linear projection. Table~\ref{tab:ablation_study} showed that removing pre-training led to a clear drop across all metrics, indicating that pre-training was essential for aligning molecular representations with language semantics. Architecture choices also affected generation quality. Replacing the fusion projector with concatenation performed worse than the plain non-pretrained model, with the largest losses on BLEU-2 and METEOR, which indicated the need for a dedicated fusion module. Removing the 3D fingerprint yielded the weakest ROUGE-L, suggesting degraded semantic fidelity. These results showed that 3D stereochemical cues and effective fusion were both necessary for accurate and chemically plausible descriptions.

% \begin{figure}[ht]
%     \centering
%     \includegraphics[width=\linewidth]{section/img/ablation_study.pdf}
%     \caption{Performance comparison of different model variants in the ablation study.}
%     \label{fig:ablation}
% \end{figure}

% \input{section/5_results_discussion}
\section{Conclusion}

We investigated dual-view molecular understanding, aligning language with molecular images and 3D structures. DeepMoLM strengthened cross-modal interaction, encoded stereochemical invariants, and improved perception with a dual-pathway encoder. It achieved competitive results across captioning, description, and property prediction, surpassing generalist baselines with valid outputs. Ablations confirmed the necessity of pre-training, fusion, and 3D fingerprints. DeepMoLM thus offers a robust framework for molecular understanding.

\section*{Appendix}
\appendix

\section{Reproducibility Statement}
We have made every effort to ensure that the results presented in this paper are reproducible. All code and datasets have been made publicly available in an anonymous repository to facilitate replication and verification. The experimental setup, including training steps, model configurations, and hardware details, is described in detail in the paper. We have also provided a full description of DeepMoLM, to assist others in reproducing our experiments. Additionally, the supplementary data, such as 2D molecule images and E3FP molecular fingerprints, are publicly available, ensuring consistent and reproducible evaluation results. We believe these measures will enable other researchers to reproduce our work and further advance the field.

\section{LLM Usage}
Large Language Models (LLMs) were used only for sentence rephrasing, grammar checking, and improving text flow. They were not involved in ideation, methodology, or experiments; all scientific content was developed by the authors. The authors take full responsibility for the manuscript, and all LLM-assisted text complies with ethical guidelines without contributing to plagiarism or misconduct.

\section*{Ethics Statement}
This work follows the IJCAI-ECAI Code of Ethics. No human or animal experiments were involved. All datasets, including PubChem and CCheBI-20, were used in compliance with guidelines and without privacy violations. No personally identifiable information was included, and no procedures posed privacy or security risks. The study was conducted with transparency and integrity.

\section*{Acknowledgments}

This work was supported by an internal grant from The Hong Kong Polytechnic University (Project No. P0051278, Jung Sun Yoo) and the General Research Fund (Project No. PolyU 15101422, Jung Sun Yoo) from the Research Grants Council of the Hong Kong Special Administrative Region, China.
\newpage
%Bibliography
\bibliographystyle{unsrt}  
\bibliography{references}

\end{document}